\documentclass[journal]{IEEEtran}
\usepackage{cite}
\usepackage{url}            
\usepackage{graphicx,graphics,color,colortbl,psfrag}
\usepackage[usenames,dvipsnames]{xcolor}
\usepackage[bookmarks=true,
			colorlinks=true,
			linkcolor=Black, 
			urlcolor=Black,  
			citecolor=Black]{hyperref}
\usepackage{url}
\usepackage{booktabs}       
\usepackage{amsfonts}       
\usepackage{nicefrac}       
\usepackage{microtype}      
\usepackage{multicol}
\usepackage{rotating}
\usepackage{amsmath}
\usepackage{mathrsfs} 
\usepackage{subcaption}
\usepackage{array,multirow}
\usepackage{lscape} 
\usepackage{array}
\newcommand{\PreserveBackslash}[1]{\let\temp=\\#1\let\\=\temp}
\newcolumntype{C}[1]{>{\PreserveBackslash\centering}p{#1}}

\usepackage{comment}
\usepackage{algorithm,algpseudocode}
\usepackage{todonotes}

\usepackage{amsthm}
\theoremstyle{definition}
\newtheorem{assumption}{Assumption}

\graphicspath{{figs/}}
\usepackage{makecell}                                                                                                        
\usepackage{colortbl}

\graphicspath{{figs/}}

\ifCLASSINFOpdf
\else
\fi

\begin{document}

\title{Multi-Task Variational Information Bottleneck}

\author{
Weizhu Qian, Bowei Chen, Yichao Zhang, Guanghui Wen, Franck Gechter\IEEEcompsocitemizethanks{Weizhu Qian is with Le laboratoire Connaissance et Intelligence Artificielle Distribu\'ees, Universit\'e Bourgogne Franche-Comt\'e, France (\href{mailto:weizhu.qian@utbm.fr
}{weizhu.qian@utbm.fr}); Bowei Chen is with the Adam Smith Business School, University of Glasgow, UK (\href{mailto:bowei.chen@glasgow.ac.uk}{bowei.chen@glasgow.ac.uk}); Yichao Zhang is with the Department of Computer Science and Technology of Tongji University, also with the Key Laboratory of Embedded System and Service Computing (Tongji University), Ministry of Education, Shanghai 200092, China (\href{mailto:yichaozhang@tongji.edu.cn}{yichaozhang@tongji.edu.cn}); Guanghui Wen is with the Department of Systems Science, School of Mathematics, Southeast University, Nanjing 210016, China, and also with the School of Engineering, RMIT University, Melbourne VIC 3000, Australia~(\href{mailto:wenguanghui@gmail.com}{wenguanghui@gmail.com}); Franck Gechter is with Le laboratoire Connaissance et Intelligence Artificielle Distribu\'ees, Universit\'e Bourgogne Franche-Comt\'e and Mosel LORIA UMR CNRS 7503, Université de Lorraine, France  (\href{mailto:franck.gechter@utbm.fr}{franck.gechter@utbm.fr}).
}
}



\maketitle

\begin{abstract}
Multi-task learning (MTL) is an important subject in machine learning and artificial intelligence. Its applications to computer vision, signal processing, and speech recognition are ubiquitous. Although this subject has attracted considerable attention recently, the performance and robustness of the existing models to different tasks have not been well balanced. This article proposes an MTL model based on the architecture of the variational information bottleneck (VIB), which can provide a more effective latent representation of the input features for the downstream tasks. Extensive observations on three public data sets under adversarial attacks show that the proposed model is competitive to the state-of-the-art algorithms concerning the prediction accuracy. Experimental results suggest that combining the VIB and the task-dependent uncertainties is a very effective way to abstract valid information from the input features for accomplishing multiple tasks.
\end{abstract}


\IEEEpeerreviewmaketitle


\section{Introduction}
\label{sec:intro}

\IEEEPARstart{M}{}ulti-task learning (MTL) is a popular sub-field of machine learning that has been widely used in many real-world applications, including natural language processing, speech recognition, autonomous driving, computer vision, ubiquitous computing, and so on~\cite{zhang2017survey, kokkinos2017ubernet, lu2014semi, liu2014multiple}. If a set of learning tasks are related to each other, they can be accomplished jointly in an MTL framework. In a typical MTL, different tasks share the same representation of input features, which is learned through minimizing the losses of the tasks simultaneously. The MTL models have been proven to be more efficient and effective than training individual tasks separately. In the MTL framework, each task contributes differently to the total loss. Therefore, how to learn a balanced representation of the input is key to the performance of the model.

\begin{figure}[t]
\centering
\includegraphics[width=1\linewidth]{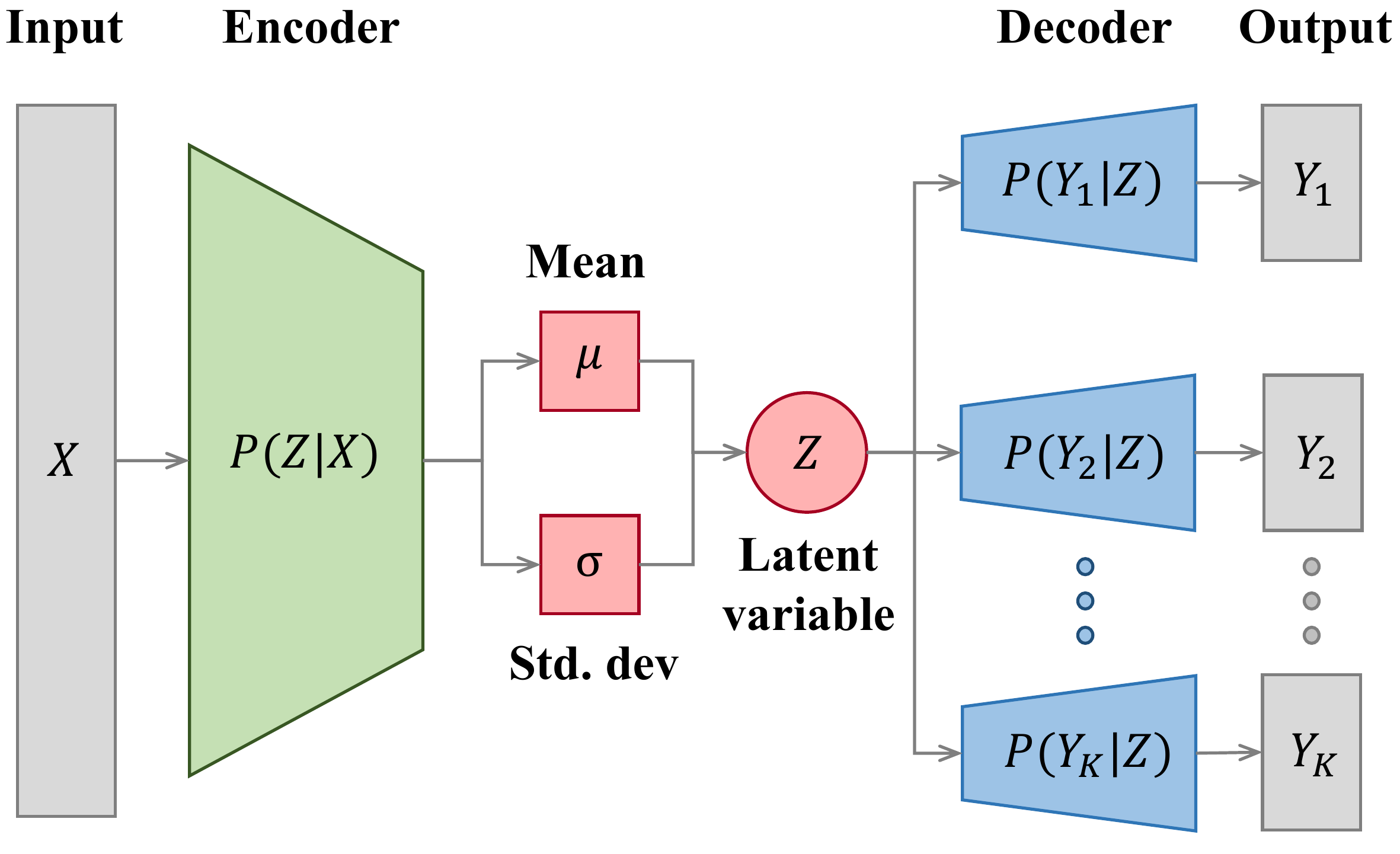}
\vspace*{-8pt}
\caption{The architecture of the MTVIB.}
\label{fig:schematic_view}
\end{figure}

The research of the MTL models can be classified into two branches~\cite{goodfellow2014explaining,zhang2017survey,ruder2017overview}. The first branch works on improving the shared latent representation while the second aims to estimate the optimal weights for the respective learning tasks. Our study in this paper covers the interests of both branches. For the former, a number of MTL models use deterministic encoders to obtain the shared latent presentation for the task-specific decoders~\cite{ruder2017overview,zhang2017survey}. The methods are typically sensitive to noises \cite{goodfellow2014explaining}. For the latter, constant relative weights are widely used to balance the downstream tasks, which may fail to find the optimal trade-offs. In order to address these issues, we propose a model based on the information bottleneck framework~\cite{tishby2000information,tishby2015deep} to learn a balanced representation of the input. The illustration of the framework is presented in Fig.~\ref{fig:schematic_view}. We use the information bottleneck to obtain the latent codes and implement it with variational inference~\cite{blei2017variational,zhang2018advances} so the model is called the \emph{variational information bottleneck} (in short VIB)~\cite{alemi2017deep}. Variational methods can provide an analytical approximation to the posterior probability of the latent representations with parameters. Notable variational methods include variational autoencoder and its variants~\cite{kingma2015variational,rezende2014stochastic,higgins2017beta,burgess2018understanding,maaloe2016auxiliary}, Bayesian neural networks~\cite{blundell2015weight}, variational dropout~\cite{kingma2015variational,molchanov2017variational} and deep variational prior~\cite{atanov2018deep}. Compared to deterministic latent representations, the variational latent representations are regularized and thereby more robust to noises such as adversarial attacks.

On the other hand, the weights of tasks in an MTL model can be set manually or estimated algorithmically \cite{kendall2018multi}. A good example is that the grid search (GS) method investigates a variety of combinations of constant weights to optimize the total loss of the tasks. Chen et al.~\cite{chen2018gradnorm} propose the gradient normalization (GradNorm) to automatically balance training in deep MTL models by dynamically tuning gradient magnitudes. Blundell et al.~\cite{blundell2015weight} define an uncertainty weighted loss (UWL), which takes a variational approximation and trains an ensemble of networks where each network has its weights drawn from the shared probability distribution. We adjust the uncertainty-based method by using the likelihoods of the predictions to weigh tasks, enabling the structure of the VIB can be applied to the MTL. With the uncertainty-weighting strategy and the VIB structure, a good trade-off among different learning tasks can be achieved, and the robustness of the MTL model can be improved.

Our study is established on an assumption that the information associated with the input data follows a latent distribution, which is also referred to as \emph{information Markov chain} in literature \cite{tishby2000information, tishby2015deep}. Let $X$ and $Z$ denote the input data and its latent representation. The output is denoted by $Y$. The information Markov chain is then $X \rightarrow Z \rightarrow Y$, which can be presented in an encoder-decoder structure. A similar idea was proposed by Achille et al. \cite{achille2018emergence}, where the information is stored in the neural network weights. In the following, we will first introduce the three assumptions of our model.

\begin{assumption}\label{ass:1}
There exists a statistic of the input $X$ that is solely sufficient enough to learn the posterior probability of $Z$. 
\end{assumption} 

This assumption was proposed by Alemi et al. \cite{alemi2017deep}. It suggests that the information in $X$ is sufficient to derive the latent distribution of $Z$. Therefore, we can simply have $P(Z|X,Y) = P(Z|X)$. The latent distribution in our proposed model can be learned by an encoder network. As opposed to the deterministic codes computed by vanilla auto-encoders~\cite{hinton1994autoencoders}, we adopt the VIB framework, and accordingly, the learned latent distribution can be seen as a disentangled representation. In an ideal situation, the latent representation is not only sufficient but also minimal. At the moment, only the task-related information is retained.

\begin{assumption}\label{ass:2}
The learned representation $Z$ is solely sufficient enough to learn the likelihood of $Y$.
\end{assumption}

This assumption was made by Achille et al.~\cite{achille2018emergence}. It indicates that the sufficiency of the latent representation is guaranteed by the decoder network. In other words, it can be represented by $P(Y|X,Z) = P(Y|Z)$. 

\begin{assumption}\label{ass:3}
If there are multiple learning tasks, they are conditionally independent and share the same representation.
\end{assumption}

This assumption was used by Kendall et al.~\cite{kendall2018multi}. It allows division of the output into different sub-outputs that can be generated by different decoders (i.e., classifiers in a classification task). Mathematically, let $Y = [Y_1, Y_2, \ldots, Y_K]$ denote multiple learning tasks, then $P (Y|Z) = \prod_{k=1}^{K} P(Y_k|Z)$. 


The rest of the paper is organized as follows. Section~\ref{sec:related_work} reviews the related literature. Section~\ref{sec:vib} introduces the mathematical preliminaries of the VIB. Section~\ref{sec:mtvib} presents our proposed model MTVIB by extending the VIB structure to multiple tasks. Section~\ref{sec:discussion} further discusses the connection of our method to other related studies. Section~\ref{sec:experiments} discusses our data sets, experimental design, settings, and results. We conclude the paper in Section~\ref{sec:conclusion}.

\section{Related Work}
\label{sec:related_work}

In recent years, the information bottleneck method has become a popular topic in machine learning. Achille et al.~\cite{achille2018information} leveraged it to develop a novel dropout method to learn the optimal representations. Dai et al.~\cite{dai2018compressing} adopted the VIB to compress the structures of neural networks. Federici et al.~\cite{federici2020learning} developed a multi-view VIB to pursue a better generalization performance. Belghazi et al.~\cite{belghazi2018mutual} applied the mutual information neural estimation to improving the VIB performance. Peng et al.~\cite{peng2019variational} proposed a variational discriminator bottleneck to improve the performance of the imitation learning, inverse reinforcement learning, and generative adversarial networks. Wu et al.~\cite{wu2020phase} studied the phase transition in the VIB structure. In a reinforcement learning task, Goyal et al.~\cite{goyal2020variational} devised an interesting variational bandwidth bottleneck to improve the generalization and reduce the computational load. 

In a sense, the VIB is similar to deep latent generative models such as variational autoencoders (VAEs)~\cite{kingma2014auto} and $\beta$-VAEs~\cite{higgins2017beta, burgess2018understanding}, since they are all used to learn stochastic latent presentations. The difference is that the VIB is a supervised learning approach to provide prediction results while VAEs and $\beta$-VAEs are unsupervised learning models used to generate new samples.

In terms of other recent MTL methods, Long et al.~\cite{long2017learning} proposed multi-linear relationship networks (MRNs) jointly learning transferable representations and task relationships by tensor normal priors over parameter tensors of multiple task-specific layers in deep convolutional networks. The MRNs can alleviate the dilemma of negative transfer in the feature layers and under-transfer in the classifier layer. Lu et al.~\cite{lu2017fully} proposed a multi-task model, which realized automatic learning of multi-task deep network architecture through a dynamic branching process. Considering and the complexity of the model, they grouped tasks at each layer of the network by the relevance among tasks. Misra et al.~\cite{misra2016cross} developed a convolutional network architecture with a special sharing unit, called “cross-stitch” unit, to learn an optimal combination of the shared and task-dependent representations for multi-task learning. Liu et al.~\cite{liu2019end} proposed an end-to-end multi-task learning model, called the multi-task attention networks (MTANs), which is composed of a single shared network containing a global feature pool and a task-specific soft attention module for each task. The authors claimed that the modules are able to automatically determine the importance of the shared features for the respective task. Therefore, the model is capable of learning both task-shared and task-specific features in a self-supervised, end-to-end manner. Xu et al.~\cite{xu2018pad} proposed a prediction-and-distillation network (PAD-Net) that leverages a set of intermediate auxiliary tasks as multi-modal input for multi-modal distillation modules to accomplish the final tasks. Rudd et al.~\cite{rudd2016moon} developed a framework, in which a mixed multiple objective with domain adaptive re-weighting of propagated loss for the facial recognition tasks. Pentina et al.~\cite{pentina2017multi} proposed an algorithm to transfer a model trained with labeled samples to accomplish the tasks without labeled samples. Gong et al.~\cite{gong2018multiobjective} discussed an algorithm to learn multi-objective time-series sequences. Feng et al.~\cite{feng2018evolutionary} investigated an evolutionary search algorithm with explicit genetic knowledge transferring across tasks via auto-encoding, which enables an effective combination of different search strategies. Zhong et al.~\cite{zhong2014learning} studied a two-stage multi-task sparse learning framework for facial recognition. Recently, how tasks interact with each other in multi-task learning likewise attracted a lot of attention \cite{standley2020tasks, zamir2018taskonomy, pal2019zero, dwivedi2019representation, achille2019task2vec}. 

As opposed to the existing methods, we combine the VIB and the UWL to create a novel MTL framework, called the \emph{multi-task variational information bottleneck} (in short MTVIB), which can achieve excellent and comparable prediction performance as the state-of-the-art models while with a better robustness to the nosiy data.

\section{Variational Information Bottleneck}
\label{sec:vib}

We introduce the related mathematical preliminaries of the VIB~\cite{alemi2017deep} in this section. The main idea of the VIB is to leverage variational inference to implement the information bottleneck. It aims to compress the input $X$ maximally, while also express the output $Y$ as much as possible via a latent stochastic representation $Z$. 

In information theory, the information bottleneck can be formulated as the following optimization problem:
\begin{align}
\label{Eq:IB_1}
\text{max}  & ~ I(Z;Y),   \\
\label{Eq:IB_2}
\text{s.t.} & ~ I(Z;X) \leq I_C,
\end{align}
where $I$ denotes the mutual information between two variables, and $I_C$ is a constant representing information constraint. From the perspective of the rate distortion, Eq.~(\ref{Eq:IB_2}) serves as an upstream representation learning task, and Eq.~(\ref{Eq:IB_1}) acts as a downstream prediction learning task. 

Analogously, our aim is to maximize the following objective function by employing the Lagrange method
\begin{align}
\label{Eq:IB_Lag}
\mathscr{L}_{\text{IB}} = I(Z;Y) - \beta \big( I(Z;X) - I_C \big),
\end{align}
where $\beta$ is a non-negative Lagrangian multiplier.

Direct computing the mutual information is challenging. Therefore, previous studies turn to variational inference for solving the information bottleneck optimization problem~\cite{alemi2017deep}, which is to minimize the following objective function
\begin{align}
\label{Eq:OP_VIB_L}
\mathscr{L}_{\text{VIB}} & =  \mathbb{E}_{\mathscr{D}} \bigg[ \mathbb{E}_{z \sim p_{\phi}(z|x)} \Big[-\text{log}\{ p_{\theta}(y|z)\} \Big] 
\nonumber \\  
& \ \ \ \ \qquad + \beta D_{\text{KL}}(p_{\phi}(z|x)|| q(z)) \bigg].  
\end{align}
where $D_{\text{KL}}$ is the Kullback-Leibler (\text{KL}) divergence, $p_{\phi}(z|x)$ is an encoder parameterized by $\phi$, $p_{\theta}(y|z)$ is a decoder parameterized by $\theta$, and $q(z)$ is an uninformative prior distribution (e.g., a standard Gaussian distribution). 

The aim of the VIB is to learn the latent representation $Z$, which is a stochastic variable and it is assumed to be sufficient to learn the output $Y$ (see Assumption~\ref{ass:2}). Meanwhile, the learned latent variable $Z$ is also a minimal representation of the input $X$ (see Assumption~\ref{ass:1}), which suggests that $Z$ contains only the task-related information. This property can be exploited to circumvent the overfitting issue and to defend adversarial attack in MTL by providing the shared stochastic codes to downstream tasks.

\section{Multi-Task Variational Information Bottleneck}
\label{sec:mtvib}

As illustrated in Fig.~\ref{fig:schematic_view}, the VIB structure plays the role of an encoder and a decoder in the MTL framework. The MTVIB has an encoder-decoder architecture with a latent distribution. The encoder network is designed to accomplish the upstream task, i.e., to attain the latent representation. The decoder network is designed to proceed the downstream tasks, i.e., output the prediction results.   

According to Assumption 3, different tasks are conditionally independent from each other. Let $I(Z;Y_k)$ be the mutual information for the downstream task $Y_k$. Inspired by the information bottleneck method, we formulate the MTL as
\begin{align}
\label{Eq:MIB_1}
\text{max}  & ~ \sum_{k=1}^K I(Z;Y_k),   \\
\label{Eq:MIB_2}
\text{s.t.} & ~ I(Z;X) \leq I_C.
\end{align}

Instead of solving the above constrained optimization problem directly, we follow the way mentioned in Eq.~(\ref{Eq:IB_Lag}) and cast the optimization problem into
\begin{align}
\label{Eq:MTIB_Lag}
\mathscr{L}_{\text{MTIB}} = \sum_{k=1}^K I(Z;Y_k) - \beta I(Z;X).
\end{align}

Direct computing the mutual information $I(Z;Y_k)$ and $I(Z;X)$ in Eq.~(\ref{Eq:MTIB_Lag}) is a challenging task. Variational inference can be employed to provide approximated solutions for them. Firstly, with Assumption 1, one can derive a variational upper bound for $I(Z;X)$ as follows

\begin{align}
\label{Eq:I_ZX}
& I(Z;X) \nonumber \\
~ = & \ D_{\text{KL}} \big(p(z,x) \| p(z) p(x) \big) \nonumber \\
~ = & \ \iint p(z,x) \text{log} \left\{ \frac{p(z,x)}{p(z) p(x)} \right\} dx dz \nonumber \\
~ = & \ \iint p(z,x) \text{log} \left\{ \frac{p(z|x) q(z)}{p(z) q(z)} \right\}  dx dz\nonumber \\
~ = & \ \iint p(z,x) \left[\text{log} \left\{ \frac{p(z|x)}{q(z)} \right\}  - \text{log} \left\{ \frac{p(z)}{q(z)} \right\} \right] dx dz \nonumber \\
~ = & \ \iint p(z|x) p(x) \text{log} \left\{ \frac{p(z|x)}{q(z)}  \right\} \nonumber \\ & ~ \qquad - \iint p(z|x) p(x) \text{log} \left\{ \frac{p(z)}{q(z)} \right\}  dx dz \nonumber \\
~ = & \ \mathbb{E}_x \left[D_{\text{KL}} \big(p(z|x)||q(z)\big)\right] - D_{\text{KL}} \Big(p(z)||q(z) \Big) \nonumber \\
~ \leq & \ \mathbb{E}_x \left[D_{\text{KL}} \big(p(z|x)||q(z)\big)\right].
\end{align}

Eq. (\ref{Eq:I_ZX}) can be estimated with Monte Carlo sampling. We draw a series of mini batch of the input $x$ and adopt the re-parameterzation method~\cite{kingma2014auto}, \cite{doersch2016tutorial} to sample $Z$ from the latent distribution according to $p(z|x)$.

\begin{align}
\label{Eq:F_Statistic}
f_\phi(x,\epsilon_z) = \mu_z(x) + \sigma_z(x) \odot \epsilon_z,
\end{align}
where $f_\phi(x,\epsilon_z)$ is a deterministic function used in encoder $p_{\phi}(z|x)$, $\odot$ is the Hadamard product, and $\epsilon_z$ is a random noise sampled from a standard diagonal Gaussian distribution $\mathcal{N}(0,\mathbb{I})$, $\mu_z$ and $\sigma_z$ are two deterministic functions to calculate the mean and variance of the latent Gaussian distribution, respectively.

Let $q(z)$ be a standard diagonal normal distribution, then $D_{\text{KL}} \big(p(z|x)||q(z)\big)$ can be calculated by 

\begin{align}
\label{Eq:D_KL}
& \ D_{\text{KL}} \big(p(z|x)||q(z)\big) \nonumber \\
= & \
\frac{1}{2} \bigg[ \text{tr}(\sigma_z^2(x)) + \mu_z(x)^\top\mu_z(x) 
- d - \text{log} \Big\{ \text{det}(\sigma_z^2(x)) \Big\} \bigg],
\end{align}
where $d$ is the dimension of the latent variable $Z$.

Then, we can derive a variational bound for $I(Z;Y_k)$. Note that the information of the relative weights for the multiple downstream tasks lie in the mutual information $I(Z;Y_k)$ per se. Accordingly, to derive a closed form of $I(Z;Y_k)$, we are required to leverage the homoscedastic uncertainties of the losses. Homoscedastic uncertainty is a quantity which stays constant for all input data and varies between different tasks instead of being the output of decoders, i.e. it is task-dependent~\cite{kendall2018multi}. Based on Assumption~3, we have
\begin{align}
\label{Eq:Prob_Weighted}
\text{log} \left\{ p(y, z)\right\}  & ~ = \sum_{k=1}^K \text{log} \left\{ p(y_k,z;\sigma_k) \right\}  \nonumber  \\
& ~ = \sum_{k=1}^K \text{log} \left\{ p(y_k|z;\sigma_k)p(z) \right\}, 
\end{align}
where $\sigma_k$ is a positive scalar used to explicitly indicate the homoscedastic uncertainty of the downstream task $k$.

With Assumption 2 and Eq.~(\ref{Eq:Prob_Weighted}), a variational lower bound of $I(Z;Y_k)$ can be derived
\begin{align}
\label{Eq:I_ZY}
& I(Z;Y_k) \nonumber \\
~ = & \ D_{\text{KL}} \big(p(y_k,z;\sigma_k) \| p(z) p(y_k) \big) \nonumber \\
~ = & \ \mathbb{E}_{p(y,z)} \bigg[ \text{log} \left\{ \frac{p(y_k,z;\sigma_k)}{p(z) p(y)} \right\} \bigg] \nonumber \\
~ = & \ \mathbb{E}_{p(y,z)} \bigg[ \text{log} \left\{ \frac{p(y_k,z;\sigma_k)}{p(z)} \right\} - \text{log}\left\{ p(y_k) \right\} \bigg] \nonumber \\
~ = & \ \mathbb{E}_{p(y,z)} \bigg[ \text{log} \left\{ \frac{p(y_k|z;\sigma_k)p(z)}{p(z)} \right\} - \text{log}\left\{ p(y_k) \right\} \bigg] \nonumber \\
~ \geq & \ \mathbb{E}_{p(y,z)} \bigg[ \text{log} \left\{ p(y_k|z;\sigma_k) \right\} \bigg].
\end{align}

Similar to $I(Z;X)$, one can likewise adopt Monte Carlo sampling to estimate $I(Z;Y_k)$. Since we have sampled a mini batch of $z$ when estimating $I(Z;X)$, the mini batch can be leveraged again to sample $y$ by a decoder neural network with a softmax function and a relative weight. And the likelihood can be measured by the cross-entropy between the prediction results and the ground truth.     

Combining Eq. (\ref{Eq:I_ZX}) and Eq. (\ref{Eq:I_ZY}), we have the following optimization objective
\begin{align}
\label{Eq:Objective}
\sum_{k=1}^K I(Z;Y_k) - \beta I(Z;X) 
\geq & \ 
\mathbb{E}_{p(y,z)} \bigg[ \sum_{k=1}^K \text{log} \left\{ p(y_k|z;\sigma_k) \right\} \bigg] \nonumber \\
& \ \ \ - \beta \mathbb{E}_x \bigg[D_{\text{KL}} \Big( p(z|x)||q(z) \Big)\bigg].
\end{align}

Maximizing the above objective is equivalent to minimizing the following loss function

\begin{align}
\label{Eq:MTVIB_Loss}
\mathscr{L}_{\text{MTVIB}} 
= & \ 
\mathbb{E}_{p(y,z)} \bigg[\sum_{k=1}^K  -\text{log} \left\{ p(y_k|z;\sigma_k) \right\} \bigg] \nonumber \\ 
& \ 
\qquad + \beta \mathbb{E}_x \bigg[D_{\text{KL}} \Big(p(z|x)||q(z)\Big)\bigg].
\end{align}

Furthermore, for classification tasks in particular, we need to construct a specific distribution. To this end, the softmax function is employed to construct a Boltzmann distribution (or Gibbs distribution). Let $f_{k}(z)$ be the output vector of the decoder network for the task $k$. 


We follow the methods used in \cite{kendall2018multi} to derive the relative weights of the downstream tasks. For the classification likelihood of class $i$ for task $k$, the element-wise standard softmax function is
\begin{align}
\label{Eq:Softmax_standard}
\text{softmax}(f_k^i(z)) = \frac{\exp \Big\{ f_k^i(z) \Big\} }{ \sum_{i=1}^I{ \exp \Big\{ f_k^i(z) \Big\} }}.
\end{align}

The unscaled element-wise softmax function with variances for element $f_k^i(z)$ can be written as:
\begin{align}
\label{Eq:Softmax_sigma}
\text{softmax}(f_k^i(z),\sigma_k) = \frac{\exp \Big\{ \frac{1}{\sigma_k^2}f_k^i(z) \Big\}}{ \sum_{i=1}^I{ \exp \Big\{ \frac{1}{\sigma_k^2}f_k^i(z) \Big\} }}. 
\end{align}

The estimating strategy is to leverage the unscaled outputs $f_k^i(z)$ to calculate Eq.~(\ref{Eq:Softmax_sigma}), where
\begin{align}
\label{Eq:Softmax_eq}
\exp\bigg\{\frac{1}{\sigma_k^2}z\bigg\} = \Big[\exp\{z\}\Big]^\frac{1}{\sigma_k^2}.
\end{align}

To normalize the likelihoods of the elements, the element-wise Softmax function with variances is rewritten as
\begin{align}
\label{Eq:Softmax_subs}
\Big[\text{softmax}(f_k^i(z)) \Big]^\frac{1}{\sigma_k^2} = \frac{\Big[\exp \Big\{f_k^i(z) \Big\}\Big]^\frac{1}{\sigma_k^2}}{\Big[\sum_{i=1}^I\exp\big\{f_k^j(z)\big\} \Big]^\frac{1}{\sigma_k^2}}, 
\end{align}


In order to compute Eq.~(\ref{Eq:Softmax_sigma}) by Eq.~(\ref{Eq:Softmax_standard}), we substitute the numerator of Eq.~(\ref{Eq:Softmax_sigma}) with Eq.~(\ref{Eq:Softmax_subs}), Eq.~(\ref{Eq:Softmax_sigma}) is rewritten as
\begin{align}
\label{Eq:Softmax_task1}
  & \ \text{softmax}(f_k^i(z),\sigma_k) \nonumber \\
= & \ 
\frac{\Big[\text{softmax}(f_k^i(z)) \Big]^\frac{1}{\sigma_k^2} \Big[ \sum_{i=1}^I\exp\big\{f_k^i(z)\big\} \Big]^\frac{1}{\sigma_k^2}  }{ \sum_{i=1}^I{ \exp \Big\{ \frac{1}{\sigma_k^2}f_k^i(z) \Big\}}}, 
\end{align}

Take the log of both sides, one has
\begin{align}
\label{Eq:Softmax_task2}
& \
\text{log} \bigg\{ \text{softmax}(f_k^i(z),\sigma_k) \bigg\} \nonumber \\
= & \ \frac{1}{\sigma_k^2} \text{log} \Big\{ \text{softmax}(f_k^i(z)) \Big\} \nonumber \\
& \ + \text{log} \left\{ \frac{\sum_{i=1}^I\exp \big\{ \frac{1}{\sigma_k^2}f^i_k(z)\big\} }{\bigg[ \sum_{i=1}^I\exp \big\{f^i_k(z) \big\} \bigg]^\frac{1}{\sigma_k^2}}\right\}.
\end{align}

We then use the following approximation~\cite{kendall2018multi} 
\begin{align}
\label{Eq:Softmax_task3}
\Bigg[ \sum_{i=1}^I \exp \big\{ f^i_k(z) \big\} \Bigg]^\frac{1}{\sigma_k^2} \approx \frac{1}{\sigma_k}\sum_{i=1}^I\exp \Big\{\frac{1}{\sigma_k^2}f^i_\theta(z)\Big\},
\end{align}
and apply it to Eq.~(\ref{Eq:Softmax_task2}). Thus, the following uncertainty weighted losses for the downstream tasks are obtained
\begin{align}
\label{Eq:softmax_task}
& \ 
\text{log} \{ p(y_k = i|z; \sigma_k) \} \nonumber \\
= & \ 
\text{log} \{ \text{softmax}(f_k^i(z),\sigma_k) \} \nonumber \\
\approx & \ 
\frac{1}{\sigma_k^2}\text{log} \{ \text{softmax}(f_k^i(z)) \} + \text{log} \left\{ \sigma_k \right\}.
\end{align}

Substituting Eq. (\ref{Eq:F_Statistic}) and Eq. (\ref{Eq:softmax_task}) into Eq. (\ref{Eq:MTVIB_Loss}), the final loss function of the proposed MTVIB can be obtained
\begin{align}
\label{Eq:MTVIB_Loss_Final}
& \mathscr{L}_{\text{MTVIB}}  \nonumber \\
= & \frac{1}{N}\sum_{n=1}^{N}\mathop{{}\mathbb{E}_{\epsilon_z \sim p(\epsilon_z)}} \Bigg[ \sum_{k=1}^{K} -\text{log} \big\{ p_{\theta_k}\big(y_n|f_{\phi}(x_n,\epsilon_z)\big) \big\} \Bigg]  
\nonumber \\  & \qquad  + \beta D_{\text{KL}}\Big( p_{\phi}\big(z|x_n)||q(z) \Big) \nonumber \\
= & \sum_{k=1}^{K} \Bigg[\frac{1}{\sigma^2_k} \mathscr{E}_{k} + \text{log} \{ \sigma_k \} \Bigg] + \beta D_{\text{KL}}\Big(p_{\phi}(z|x)||q(z)\Big). 
\end{align}
where $N$ is the size of mini batch for Monte Carlo sampling and $\mathscr{E}_{k}$ is the cross-entropy for task $k$.

Similar to the VIB, to acquire the prediction outputs, we first sample the latent variable $z$ using the encoder $p_{\phi}(z|x)$, then $z$ will be used as the input for the respective decoders  decoders $p_{\theta_k}(y_k|z)$ whose outputs are our model's predictions.

\section{Discussion}
\label{sec:discussion}


Following the common practice ~\cite{sener2018multi},~\cite{lin2019pareto}, we interpret the MTL task into a multi-objective optimization problem. The optimization objective and its constraints in our method are both based on the mutual information $I(Z;X)$ and $I(Z;Y)$. By incorporating the VIB structure~\cite{alemi2017deep}, the shared latent representation $Z$ can be maximally compressed from the input while also sufficiently represents the targets in the upstream task and the downstream task, respectively. 

The upstream task aims to compress the input features evaluated by the mutual information $I(Z;X)$ via a variational lower bound $ \mathbb{E}_x \left[D_{\text{KL}} \big(p(z|x)||q(z)\big)\right]$. The downstream task aims to estimate the mutual information $I(Z;Y)$ which can be cast into a negative log-likelihood $\mathbb{E}_{p(z,y)} \big [ -\text{log} \left\{ p(y|z) \right\} \big]$. Since $\mathbb{E}_x \left[D_{\text{KL}} \big(p(z|x)||q(z)\big)\right]$ can be regarded as the compression rate $R$, and $\mathbb{E}_{p(z,y)} \big [\text{log} \left\{ p(y|z) \right\} \big]$ can be regarded as the distortion $D$, the loss function of our model can be simply rewritten as 
\begin{align}
\label{Eq:rate_distortion}
-D + \beta R.
\end{align}

Therefore, the trade-off between the upstream task and the downstream task in our approach can also be framed as the \emph{rate-distortion trade-off}~\cite{tschannen2018recent}. Additionally, our model is in line with the minimum description length principle and provides a general solution to the over-fitting problem in the MTL~\cite{rissanen1978modeling,grunwald2007minimum}. 


Generally speaking, most existing MTL models share two major limitations: (i) the model's performance is sensitive to the relative weights assigned to the learning tasks; and (ii) the training process is computationally expensive. To tackle both issues, Blundell et al.~\cite{blundell2015weight} proposed the UWL. The model estimates the task-dependent uncertainties and uses them to assign the relative weights to the respective losses of the tasks. Whereas, the UWL still follows a conventional encoding setting. In other words, different tasks share the representation in a deterministic way, and the model therefore may not be robust enough to a small noise in the input. To tackle this issue, in our proposed method, we adopt the information bottleneck structure to generate the shared stochastic codes as the input of the decoders, which makes our MTL solution more resilient to noise.

\section{Experiments}
\label{sec:experiments}

In this section, the used data sets are firstly introduced. We then provide the details of experimental design, including adversarial examples generation, benchmark models and model training settings. The experimental results are finally presented and discussed.

\subsection{Data Sets}

To analyze the performance of the proposed MTVIB, we perform experiments across the following three publicly available data sets:
\begin{itemize}
    \item \textbf{Two-task classifications}: the MultiMNIST data set and the MultiFashion data set~\cite{lin2019pareto, xu2019multi, eslami2016attend};
    \item \textbf{Four-task classifications}: the MTFL data set~\cite{zhang2014facial, ranjan2017hyperface, li2020deep}.
\end{itemize}
These data sets have been used in the existing MTL literature. The MultiMNIST data set is created from the MNIST data set~\cite{lecun1998gradient} and the MultiFashion data set is created from the FashionMNIST data set~\cite{xiao2017fashion}. Images in both the MultiMNIST data set and the MultiFashion data set are with resolution size of $36 \times 36$ while images in the MTFL data set are with the resolution size of $150 \times 150$. We randomly sample images from these data sets: for the MultiMNIST data set and the MultiFashion data set, 120,000 images are sampled into the training set, and 10,000 images are sampled into the test set; for the MTFL data set, 9,000 images are sampled into the training set, and 1,000 images are sampled into the test set.

\subsection{Experimental Design}

\begin{figure}[tp]
\parbox[b]{1\linewidth}{\includegraphics[width=1\hsize]{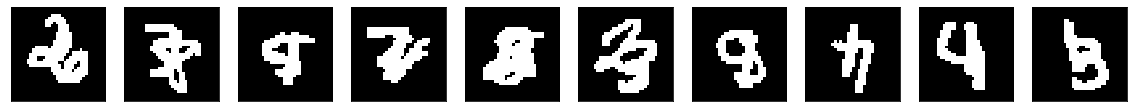}} \\
\parbox[b]{1\linewidth}{\includegraphics[width=1\hsize]{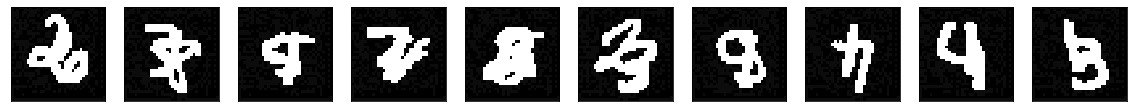}} \\
\parbox[b]{1\linewidth}{\includegraphics[width=1\hsize]{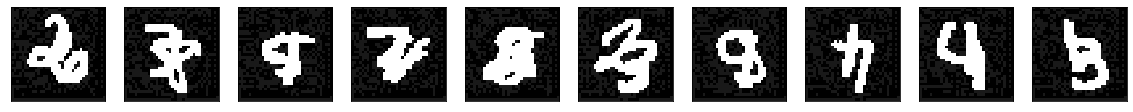}} \\
\parbox[b]{1\linewidth}{\includegraphics[width=1\hsize]{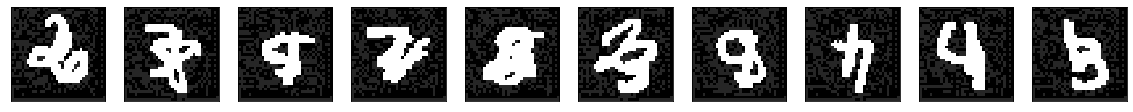}} \\
\parbox[b]{1\linewidth}{\includegraphics[width=1\hsize]{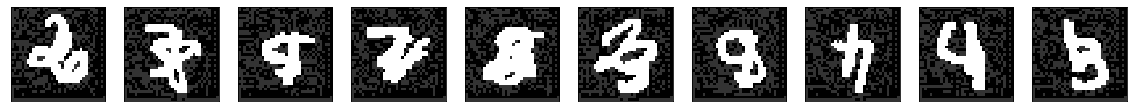}} \\
\parbox[b]{1\linewidth}{\includegraphics[width=1\hsize]{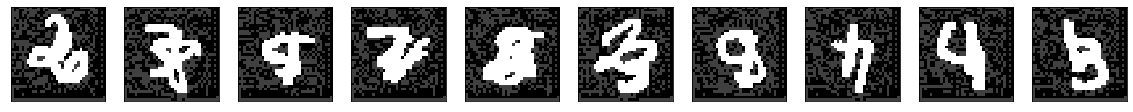}} \\
\parbox[b]{1\linewidth}{\includegraphics[width=1\hsize]{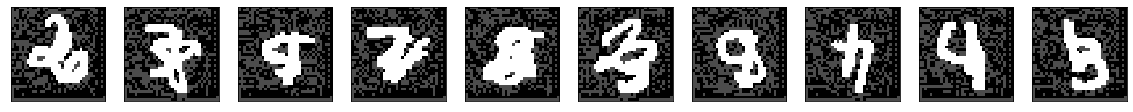}}\\
\vspace*{-8pt}
\caption{Example images of the MultiMNIST data set: original samples (the first row), adversarial samples with $\eta=0.05$ (the second row), $\eta=0.1$ (the third row), $\eta=0.15$ (the fourth row), $\eta=0.2$ (the fifth row), $\eta=0.25$ (the sixth row), and $\eta=0.3$ (the seventh row).}
\label{fig:data_sample_multimnist}
\end{figure}

\begin{figure}[tp]
\parbox[b]{1\linewidth}{\includegraphics[width=1\hsize]{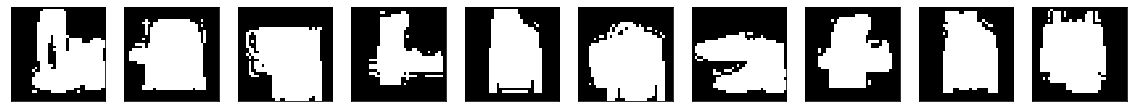}} \\
\parbox[b]{1\linewidth}{\includegraphics[width=1\hsize]{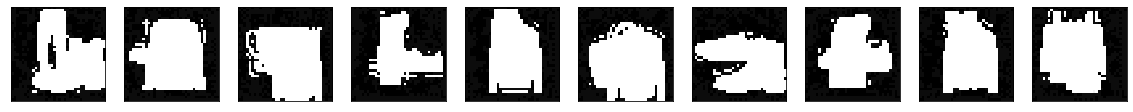}} \\
\parbox[b]{1\linewidth}{\includegraphics[width=1\hsize]{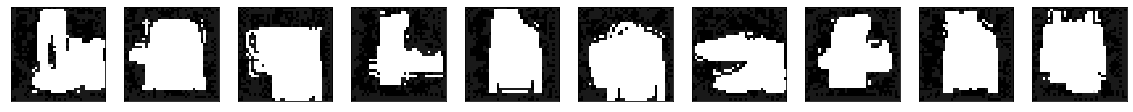}} \\
\parbox[b]{1\linewidth}{\includegraphics[width=1\hsize]{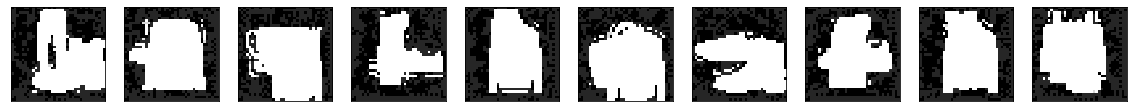}} \\
\parbox[b]{1\linewidth}{\includegraphics[width=1\hsize]{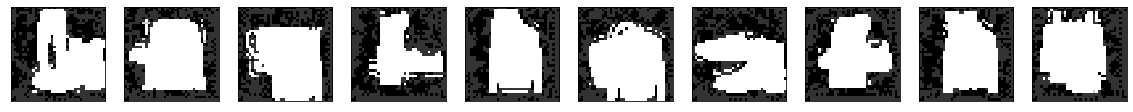}} \\
\parbox[b]{1\linewidth}{\includegraphics[width=1\hsize]{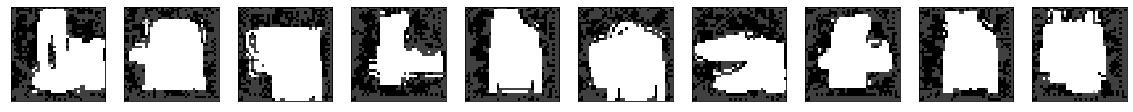}} \\
\parbox[b]{1\linewidth}{\includegraphics[width=1\hsize]{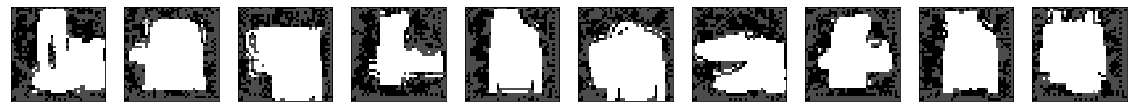}} \\
\vspace*{-8pt}
\caption{Example images of the MultiFashion data set: original samples (the first row), adversarial samples with $\eta=0.05$ (the second row), $\eta=0.1$ (the third row), $\eta=0.15$ (the fourth row), $\eta=0.2$ (the fifth row), $\eta=0.25$ (the sixth row), and $\eta=0.3$ (the seventh row).}
\label{fig:data_sample_multifashion}
\end{figure}

\begin{figure}[tp]
\parbox[b]{1\linewidth}{\includegraphics[width=1\hsize]{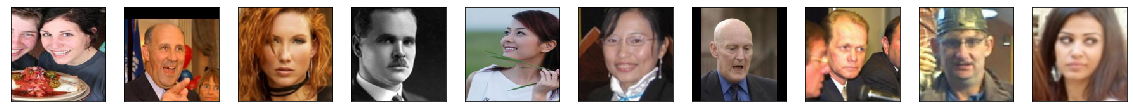}} \\
\parbox[b]{1\linewidth}{\includegraphics[width=1\hsize]{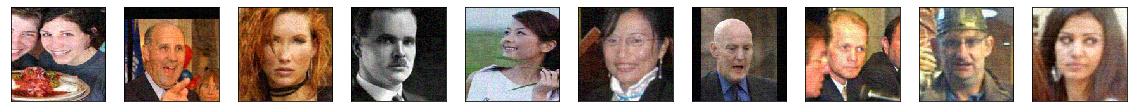}} \\
\parbox[b]{1\linewidth}{\includegraphics[width=1\hsize]{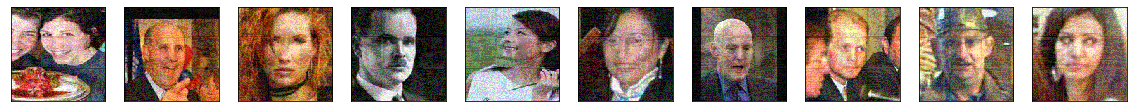}} \\
\parbox[b]{1\linewidth}{\includegraphics[width=1\hsize]{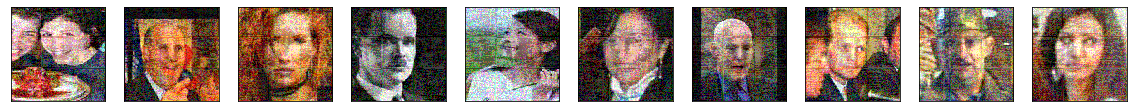}} 
\parbox[b]{1\linewidth}{\includegraphics[width=1\hsize]{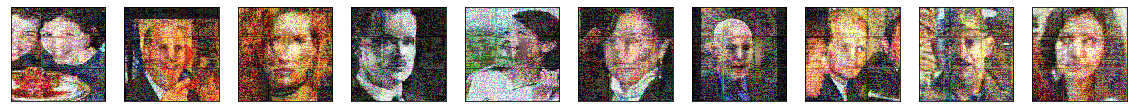}} \\
\parbox[b]{1\linewidth}{\includegraphics[width=1\hsize]{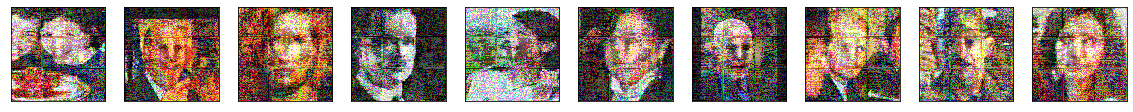}} \\
\parbox[b]{1\linewidth}{\includegraphics[width=1\hsize]{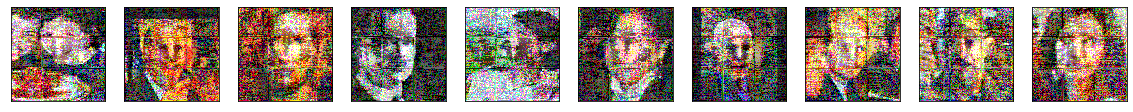}}
\vspace*{-8pt}
\caption{Example images of the MTFL data set: original samples (the first row), adversarial samples with $\eta=0.05$ (the second row), $\eta=0.1$ (the third row), $\eta=0.15$ (the fourth row), $\eta=0.2$ (the fifth row), $\eta=0.25$ (the sixth row), and $\eta=0.3$ (the seventh row).}
\label{fig:data_sample_mtfl}
\end{figure}

Deep learning models are susceptible to adversarial examples. Feeding a model with maliciously perturbed test data is widely used to examine the model's robustness. In the experiments, we use the fast gradient sign method (FGSM)~\cite{goodfellow2014explaining} to generate the adversarial examples. Specifically, let $\widetilde{x}$ denote an adversarial example, it can be obtained from its original image $x$ as follows:
\begin{align}
\label{Eq:adversarial}
\widetilde{x} = x + \eta \cdot \text{sign}\big(\nabla_x \mathscr{L}(w;x;y)\big)
\end{align}
where $\mathscr{L}$ is the loss function, $w$ denotes the parameters, and $\eta$ is a positive scalar which represents the magnitude of the perturbation. Here we test a series of values of $\eta$, taking $0.05$, $0.1$, $0.15$, $0.2$, $0.25$, $0.3$ to generate adversarial examples. For illustrative purposes, the original image examples and their adversarial examples for the used data sets are presented in Figs.~\ref{fig:data_sample_multimnist}-\ref{fig:data_sample_mtfl}, respectively. It is not difficult to see that the larger the value of $\eta$ more noises have been included in the created adversarial examples. 

The proposed MTVIB is also compared with a range of related models, including the grid search (GS)~\cite{lin2019pareto}, the uncertainty weighted losses (UWL)~\cite{kendall2018multi}, the single-task learning (STL), and the single-task variational information bottleneck (STVIB). The GS and the UWL are the state-of-the-art MTL models while the STL and the STVIB are the single task models trained separately for each learning task. We follow the experimental settings of Zhang et al.~\cite{zhang2016understanding} and use the Small AlexNet as the base neural network architecture for the STL, the GS, and the UWL. In the STVIB and the MTVIB, the dimension of the latent representation is set to $256$ and $\beta$ is set to $10^{-3}$. The default learning rate in model training is $10^{-4}$; the optimizer is the Adam~\cite{kingma2014adam}; the training epoch is $200$; and the minibatch size is $200$. For further implementation details of the examined models, please refer to Tables~\ref{tab:setting_two_tasks}~\ref{tab:setting_four_tasks}. It should be noted that: (i) the single-task models use the same architecture as the multi-task models except the number of decoders; (ii) the decoders with the same output dimensions have the same architecture.

\begin{figure*}[tp]
\begin{minipage}[t]{1\textwidth}
\begin{table}[H]
\caption{Implementation details on the MultiMNIST data set and the MultiFashion data set with: (a) the Small AlexNet model for single-task learning; (b) the Small AlexNet-adapted model for multi-task learning; (c) the STVIB model; and (d) the MTVIVB model.}
\label{tab:setting_two_tasks}
\begin{subtable}[t]{0.5\textwidth}
\centering
\caption{(a)}
\label{Tab:STL_2}
\begin{tabular}{|C{6.8cm}|}\hline
Input: 36 $\times$ 36  \\\hline
Conv 64 (Kernel: 3$\times$3; Stride: 1) + BN + ReLU \\\hline
MaxPool 2 $\times$ 2\\\hline
Conv 64 (Kernel: 3$\times$3; Stride: 1) + BN + ReLU \\\hline
MaxPool 2 $\times$ 2 \\ \hline
FC 3136 $\times$ 384 + BN + ReLU \\\hline
FC 384 $\times$ 192 + BN + ReLU \\\hline
FC 192 $\times$ 10\\\hline
softmax \\\hline
\end{tabular}
\end{subtable}
\begin{subtable}[t]{0.5\textwidth}
\centering
\caption{(b)}
\label{Tab:MTL_2}
\begin{tabular}{|C{3.55cm}|C{3.55cm}|}
\hline
\multicolumn{2}{|c|}{Input: 36 $\times$ 36} \\\hline
\multicolumn{2}{|c|}{Conv 64 (Kernel: 3$\times$3; Stride: 1) + BN + ReLU}\\ \hline
\multicolumn{2}{|c|}{MaxPool 2 $\times$ 2} \\ \hline
\multicolumn{2}{|c|}{Conv 64 (Kernel: 3$\times$3; Stride: 1) + BN + ReLU}\\ \hline
\multicolumn{2}{|c|}{MaxPool 2 $\times$ 2} \\ \hline
\multicolumn{2}{|c|}{FC 3136 $\times$ 384 + BN + ReLU} \\ 
\cline{1-2}
FC 384 $\times$ 192 + BN + ReLU & FC 384 $\times$ 192 + BN + ReLU \\
\hline
FC 192 $\times$ 10 & FC 192 $\times$ 10 \\
\hline
softmax & softmax \\
\hline
\end{tabular}
\end{subtable}

\vspace*{15pt}

\begin{subtable}[t]{0.5\textwidth}
\centering
\caption{(c)}
\label{Tab:STVIB_2}
\begin{tabular}{|C{3.2cm}|C{3.2cm}|}\hline
\multicolumn{2}{|c|}{Input: 36 $\times$ 36}  \\\hline
\multicolumn{2}{|c|}{Conv 64 (Kernel: 3$\times$3; Stride: 1) + BN + ReLU} \\\hline
\multicolumn{2}{|c|}{MaxPool 2 $\times$ 2}\\\hline
\multicolumn{2}{|c|}{Conv 64 (Kernel: 3$\times$3; Stride: 1) + BN + ReLU} \\\hline
\multicolumn{2}{|c|}{MaxPool 2 $\times$ 2} \\ \hline
\multicolumn{2}{|c|}{FC 3136 $\times$ 1024 + BN + ReLU} \\ \hline
\multicolumn{2}{|c|}{FC 1024 $\times$ 1024 + BN + ReLU} \\ \hline 
\multicolumn{2}{|c|}{FC 1024 $\times$ 1024 + BN + ReLU} \\ 
\cline{1-2}
FC 1024 $\times$ 256 + BN + ReLU & FC 1024 $\times$ 256 + BN + ReLU \\\hline 
\multicolumn{2}{|c|}{Latent dimension: 256; $\beta$: $10^{-3}$}  \\\hline
\multicolumn{2}{|c|}{FC 256 $\times$ 384 + BN + ReLU} \\\hline
\multicolumn{2}{|c|}{FC 384 $\times$ 192 + BN + ReLU} \\\hline
\multicolumn{2}{|c|}{FC 192 $\times$ 10} \\\hline
\multicolumn{2}{|c|}{softmax }\\\hline
\end{tabular}
\end{subtable}
\begin{subtable}[t]{0.5\textwidth}
\centering
\caption{(d)}
\label{Tab:MTVIB_2}
\begin{tabular}{|C{3.55cm}|C{3.55cm}|}
\hline
\multicolumn{2}{|c|}{Input: 36 $\times$ 36} \\\hline
\multicolumn{2}{|c|}{Conv 64 (Kernel: 3$\times$3; Stride: 1) + BN + ReLU}\\ \hline
\multicolumn{2}{|c|}{MaxPool 2 $\times$ 2} \\ \hline
\multicolumn{2}{|c|}{Conv 64 (Kernel: 3$\times$3; Stride: 1) + BN + ReLU}\\ \hline
\multicolumn{2}{|c|}{MaxPool 2 $\times$ 2} \\ \hline
\multicolumn{2}{|c|}{FC 3136 $\times$ 1024 + BN + ReLU} \\ \hline
\multicolumn{2}{|c|}{FC 1024 $\times$ 1024 + BN + ReLU} \\ \hline 
\multicolumn{2}{|c|}{FC 1024 $\times$ 1024 + BN + ReLU} \\ 
\cline{1-2}
FC 1024 $\times$ 256 + BN + ReLU & FC 1024 $\times$ 256 + BN + ReLU \\\hline 
\multicolumn{2}{|c|}{Latent dimension: 256; $\beta$: $10^{-3}$}  \\
\cline{1-2}
FC 256 $\times$ 384 + BN + ReLU & FC 256 $\times$ 384 + BN + ReLU \\
\hline
FC 384 $\times$ 192 & FC 384 $\times$ 192 \\
\hline
FC 192 $\times$ 10 & FC 192 $\times$ 10  \\
\hline
softmax & softmax \\
\hline
\end{tabular}
\end{subtable}
\end{table}

\begin{table}[H]
\caption{Implementation details on the MTFL data set with: (a) the Small AlexNet model for single-task learning; (b) the Small AlexNet-adapted model for multi-task learning; (c) the STVIB model; and (d) the MTVIVB model.}
\label{tab:setting_four_tasks}
\begin{subtable}[t]{0.5\textwidth}
\centering
\caption{(a)}
\label{Tab:STL_4}
\begin{tabular}{|C{6.8cm}|}\hline
Input: 150 $\times$ 150  \\\hline
Conv 64 (Kernel: 5$\times$5; Stride: 2) + BN + ReLU \\\hline
MaxPool 2 $\times$ 2\\\hline
Conv 64 (Kernel: 5$\times$5; Stride: 2) + BN + ReLU \\\hline
MaxPool 2 $\times$ 2 \\ \hline
FC 4096 $\times$ 384 + BN + ReLU \\\hline
FC 384 $\times$ 192 + BN + ReLU \\\hline
FC 192 $\times$ 5/2/2/2 \\\hline
softmax \\\hline
\end{tabular}
\end{subtable}
\begin{subtable}[t]{0.5\textwidth}
\centering
\caption{(b)}
\label{Tab:MTL_4}
\begin{tabular}{|C{3.2cm}|C{3.2cm}|C{0.3cm}|}
\hline
\multicolumn{3}{|c|}{Input: 150 $\times$ 150} \\\hline
\multicolumn{3}{|c|}{Conv 64 (Kernel: 5$\times$5; Stride: 2) + BN + ReLU}\\ \hline
\multicolumn{3}{|c|}{MaxPool 2 $\times$ 2} \\ \hline
\multicolumn{3}{|c|}{Conv 64 (Kernel: 5$\times$5; Stride: 2) + BN + ReLU}\\ \hline
\multicolumn{3}{|c|}{MaxPool 2 $\times$ 2} \\ \hline
\multicolumn{3}{|c|}{FC 4096 $\times$ 384 + BN + ReLU} \\ 
\cline{1-3}
FC 384 $\times$ 192 + BN + ReLU & FC 384 $\times$ 192 + BN + ReLU & \multirow{3}{*}{$\times$ 3} \\
\cline{1-2}
FC 192 $\times$ 5 & FC 192 $\times$ 2 &\\
\cline{1-2}
softmax & softmax &\\
\hline
\end{tabular}
\end{subtable}

\vspace*{15pt}

\begin{subtable}[t]{0.5\textwidth}
\centering
\caption{(c)}
\label{Tab:STVIB_4}
\begin{tabular}{|C{3.2cm}|C{3.2cm}|}\hline
\multicolumn{2}{|c|}{Input: 150 $\times$ 150}  \\\hline
\multicolumn{2}{|c|}{Conv 64 (Kernel: 5$\times$5; Stride: 2) + BN + ReLU} \\\hline
\multicolumn{2}{|c|}{MaxPool 2 $\times$ 2}\\\hline
\multicolumn{2}{|c|}{Conv 64 (Kernel: 5$\times$5; Stride: 2) + BN + ReLU} \\\hline
\multicolumn{2}{|c|}{MaxPool 2 $\times$ 2} \\ \hline
\multicolumn{2}{|c|}{FC 4096 $\times$ 1024 + BN + ReLU} \\ \hline
\multicolumn{2}{|c|}{FC 1024 $\times$ 1024 + BN + ReLU} \\ \hline 
\multicolumn{2}{|c|}{FC 1024 $\times$ 1024 + BN + ReLU} \\ 
\cline{1-2}
FC 1024 $\times$ 256 + BN + ReLU & FC 1024 $\times$ 256 + BN + ReLU \\\hline 
\multicolumn{2}{|c|}{Latent dimension: 256; $\beta$: $10^{-3}$}  \\\hline
\multicolumn{2}{|c|}{FC 256 $\times$ 384 + BN + ReLU} \\\hline
\multicolumn{2}{|c|}{FC 384 $\times$ 192 + BN + ReLU} \\\hline
\multicolumn{2}{|c|}{FC 192 $\times$ 5/2/2/2 } \\\hline
\multicolumn{2}{|c|}{softmax }\\\hline
\end{tabular}
\end{subtable}
\begin{subtable}[t]{0.5\textwidth}
\centering
\caption{(d)}
\label{Tab:MTVIB_4}
\begin{tabular}{|C{3.2cm}|C{3.2cm}|C{0.3cm}|}
\hline
\multicolumn{3}{|c|}{Input: 150 $\times$ 150} \\\hline
\multicolumn{3}{|c|}{Conv 64 (Kernel: 5$\times$5; Stride: 2) + BN + ReLU}\\ \hline
\multicolumn{3}{|c|}{MaxPool 2 $\times$ 2} \\ \hline
\multicolumn{3}{|c|}{Conv 64 (Kernel: 5$\times$5; Stride: 2) + BN + ReLU}\\ \hline
\multicolumn{3}{|c|}{MaxPool 2 $\times$ 2} \\ \hline
\multicolumn{3}{|c|}{FC 4096 $\times$ 1024 + BN + ReLU} \\ \hline
\multicolumn{3}{|c|}{FC 1024 $\times$ 1024 + BN + ReLU} \\ \hline 
\multicolumn{3}{|c|}{FC 1024 $\times$ 1024 + BN + ReLU} \\ \hline
{\makecell{FC 1024 $\times$ 256 + BN +\\ ReLU}} & \multicolumn{2}{|c|}{\makecell{FC 1024 $\times$ 256 + BN +\\ ReLU }}\\\hline 
\multicolumn{3}{|c|}{Latent dimension: 256; $\beta$: $10^{-3}$}  \\
\cline{1-3}
FC 256 $\times$ 384 + BN + ReLU & FC 256 $\times$ 384 + BN + ReLU & \multirow{4}{*}{$\times$ 3} \\
\cline{1-2}
FC 384 $\times$ 192 & FC 384 $\times$ 192 & \\
\cline{1-2}
FC 192 $\times$ 5 & FC 192 $\times$ 2 & \\
\cline{1-2}
softmax & softmax & \\
\hline
\end{tabular}
\end{subtable}
\end{table}
\end{minipage}
\end{figure*}

\subsection{Results}

The experimental results validate our proposed MTVIB. The evaluation can be summarized in terms of two aspects. First, the MTVIB achieves excellent and comparable performance as the benchmarked models if there is no noise in data. Second, the MTVIB significantly outperforms the benchmarked models when classification problems containing noise.  

In Tables~\ref{tab:two_task1}-\ref{tab:four_task}, we present and compare the prediction performance of all models for each classification task on the test set of the used three data sets, respectively. When there is no adversarial attack, the MTVIB achieves 94.3\% average accuracy on the MultiMNIST data set (94.9\% for the first task and 93.7\% for the second task);  85.3\% average accuracy on the MultiFashion data set (86.2\% for the first task and 84.4\% for the second task); and 74.2\% average accuracy on the MTFL data set (76.4\% for the first task, 62.4\% for the second task, 79.0\% for the third task and 78.9\% for the fourth task). It is the best-performing MTL model on the MTFL data set while ranks the second on the MultiMNIST data set and the third on the MultiFashion data set. The performance of the MTVIB is on par with the benchmarked multi-task and single-task models, and the differences in models' performance are small (roughly 0.5\%). When the level of adversarial attack increases, the performance of the MTVIB decays much slower than other models across the examined data sets. This shows that our model is more robust to the noisy data. 

\begin{figure*}[tp]
\centering
\includegraphics[width=1\linewidth]{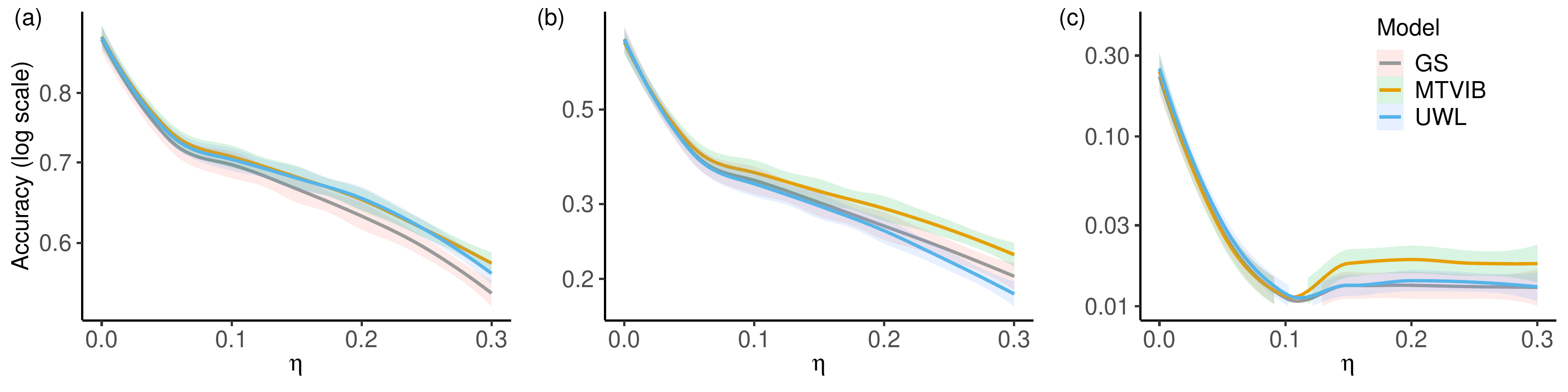}
\vspace*{-8pt}
\caption{The performance of the joint (test) accuracy of the MTL models on: (a) the MultiMNIST data set; (b) the MultiFashion data set; and (c) the MTFL data set.}
\label{fig:joint_performance}
\end{figure*}

To further examine model robustness, the joint predication accuracy of the multiple tasks by the MTL models are presented in Fig.~\ref{fig:joint_performance}, where the accuracy is calculated based on the examples whose all tasks are correctly classified. Therefore, the absolute value of the joint accuracy itself can be small but the metric gives a comparison of the models on the quality of the classification. For the MultiMNIST data set, the MTVIB and the UWL achieve comparable performance, and both models are ahead of the GS. However, the MTVIB significantly outperforms the UWL and the GS on the MultiFashion data set and the MTFL data set. Thus, Fig.~\ref{fig:joint_performance} further shows that the MTVIB has the best quality of the robustness for multi-task classifications.


\begin{table}[tp]
\centering
\caption{Summary of model performance (test accuracy) with the MultiMNIST data set. The MTL models are highlighted colored gray and the best-performing model is highlighted in bold.}
\label{tab:two_task1}
\begin{tabular}{c|c|c|c|>{\columncolor[gray]{0.8}}c|>{\columncolor[gray]{0.8}}c|>{\columncolor[gray]{0.8}}c}
\toprule
Task & $\epsilon$ & STL & STVIB  & GS     & UWL    & MTVIB\\ 
\toprule
\multirow{7}{*}{1} 
& 0  & 94.8\% & 94.9\% & 94.7\% & \textbf{95.0\%} & 94.9\%\\ \cline{2-7}
& 0.05 & 86.1\% & 86.5\% & 85.3\% & 85.6\% & \textbf{86.0\%}\\\cline{2-7}
& 0.10 & 85.4\% & 85.2\% & 84.7\% & 76.3\% & \textbf{84.7\%}\\\cline{2-7}
&0.15  & 84.0\% & 83.2\% & 83.2\% & 66.9\% & \textbf{83.4\%}\\\cline{2-7}
&0.20  & 82.5\% & 81.1\% & 81.4\% & 57.6\% & \textbf{81.6\%}\\\cline{2-7}
&0.25  & 80.2\% & 78.5\% & \textbf{79.3\%} & 48.2\% & 79.1\%\\\cline{2-7}
&0.30  & 77.1\% & 75.5\% & 76.4\% & 38.9\% & \textbf{76.8\%}\\
\toprule
\multirow{7}{*}{2} 
&0     &93.6\%   &93.7\%  &\textbf{93.7\%} &\textbf{93.7\%} &\textbf{93.7\%} \\ \cline{2-7}
&0.05  &81.9\%   &82.9\%  &80.1\% &81.1\% &\textbf{81.5\%}\\\cline{2-7}
&0.10  &80.3\%   &81.5\%  &79.3\% &80.2\% &\textbf{80.5\%}\\\cline{2-7}
&0.15  &79.0\%   &80.0\%  &77.3\% &78.8\% &\textbf{79.0\%}\\\cline{2-7}
&0.20  &76.6\%   &78.0\%  &75.1\% &77.1\% &\textbf{77.3\%}\\\cline{2-7}
&0.25  &73.7\%   &75.9\%  &72.6\% &74.2\% &\textbf{74.7\%}\\\cline{2-7}
&0.30  &70.0\%   &73.0\%  &69.0\% &71.0\% &\textbf{72.2\%}\\
\bottomrule
\end{tabular}
\vspace*{10pt}
\caption{Summary of model performance (test accuracy) with the MultiFashion data set. The MTL models are highlighted colored gray and the best-performing model is highlighted in bold.}
\label{tab:two_task2}
\begin{tabular}{c|c|c|c|>{\columncolor[gray]{0.8}}c|>{\columncolor[gray]{0.8}}c|>{\columncolor[gray]{0.8}}c}
\toprule
Task & $\epsilon$ & STL & STVIB  & GS     & UWL    & MTVIB\\ 
\toprule
\multirow{7}{*}{1}
&0     &87.1\%   &86.7\% &\textbf{86.8\%} &86.7\% &86.2\% \\  \cline{2-7}
&0.05  &59.1\%   &60.1\% &57.1\% &57.5\% &\textbf{58.2\%} \\ \cline{2-7}
&0.10  &57.3\%   &57.7\% &54.8\% &54.9\% &\textbf{56.3\%} \\ \cline{2-7}
&0.15  &54.9\%   &55.0\% &51.9\% &51.6\% &\textbf{53.7\%} \\ \cline{2-7}
&0.20  &51.7\%   &52.4\% &48.6\% &48.3\% &\textbf{51.0\%} \\ \cline{2-7}
&0.25  &48.4\%   &49.7\% &45.4\% &44.7\% &\textbf{48.4\%} \\ \cline{2-7}
&0.30  &449.\%   &46.7\% &41.9\% &40.5\% &\textbf{45.2\%} \\ 
\toprule
\multirow{7}{*}{2}
&0       &86.0\% &85.3\% &\textbf{85.6\%} &85.3\% &84.4\%\\ \cline{2-7}
&0.05    &60.8\% &60.1\% &\textbf{57.2\%} &56.3\% &56.4\%\\ \cline{2-7}
&0.10    &59.2\% &57.6\% &\textbf{55.9\%} &54.5\% &55.0\%\\ \cline{2-7}
&0.15    &56.5\% &54.7\% &\textbf{53.5\%} &52.5\% &52.8\%\\ \cline{2-7}
&0.20    &53.2\% &51.9\% &50.6\% &49.7\% &\textbf{50.9\%}\\ \cline{2-7}
&0.25    &50.3\% &49.2\% &48.3\% &46.0\% &\textbf{49.0\%}\\ \cline{2-7}
&0.30    &46.1\% &46.4\% &45.8\% &43.6\% &\textbf{46.4\%}\\
\bottomrule
\end{tabular}
\vspace*{-8pt}
\end{table}

\begin{table}[tp]
\centering
\caption{Summary of model performance (test accuracy) with the MTFL data set. The MTL models are highlighted colored gray and the best-performing model is highlighted in bold.}
\label{tab:four_task}
\begin{tabular}{c|c|c|c|>{\columncolor[gray]{0.8}}c|>{\columncolor[gray]{0.8}}c|>{\columncolor[gray]{0.8}}c}
\toprule
Task & $\epsilon$ & STL & STVIB  & GS     & UWL    & MTVIB\\ 
\toprule
\multirow{7}{*}{1} 
&0     &74.6\%  &74.3\% &72.4\%  &74.0\% &\textbf{76.4\%}\\ \cline{2-7}
&0.05  &21.1\%  &12.4\% &31.5 \% &32.1\% &\textbf{34.4\%}\\ \cline{2-7}
&0.10  &20.4\%  &12.2\% &27.4\%  &28.8\% &\textbf{34.6\%}\\ \cline{2-7}
&0.15  &20.5\%  &9.8\%  &26.7\%  &27.2\% &\textbf{32.9\%}\\ \cline{2-7}
&0.20  &20.2\%  &12.2\% &25.7\%  &26.4\% &\textbf{32.8\%}\\ \cline{2-7} 
&0.25  &20.3\%  &10.4\% &25.4\%  &26.0\% &\textbf{32.6\%}\\ \cline{2-7}
&0.30  &20.4\%  &9.6\% &24.8\%  &25.8\% &\textbf{34.4\%}\\
\toprule
\multirow{7}{*}{2} 
&0     &63.5\%  &61.3\%  &\textbf{63.3\%} &62.7\% &62.4\%\\ \cline{2-7}
&0.05  &20.3\%  &3.3\%  &21.7\% &\textbf{22.8\%} &22.6\%\\ \cline{2-7}
&0.10  &20.2\%  &3.5\%  &19.8\% &19.9\% &\textbf{22.7\%}\\ \cline{2-7}
&0.15  &20.3\%  &3.3\%  &19.6\% &19.9\% &\textbf{21.3\%}\\ \cline{2-7}
&0.20  &20.4\%  &3.4\%  &18.5\% &19.4\% &\textbf{22.6\%}\\ \cline{2-7}
&0.25  &20.3\%  &4.8\%  &18.0\% &18.9\% &\textbf{22.7\%}\\ \cline{2-7}
&0.30  &20.4\%  &5.2\%  &18.0\% &18.7\% &\textbf{22.1\%}\\
\toprule
\multirow{7}{*}{3} 
&0     &82.7\%  &81.1\%  &80.2\%  &\textbf{81.8\%} &79.0\%\\ \cline{2-7}
&0.05  &38.3\%  &63.8\%  &33.2\%  &\textbf{38.0\%} &35.2\%\\ \cline{2-7}
&0.10  &34.0\%  &64.2\%  &32.7\%  &\textbf{37.9\%} &35.6\%\\ \cline{2-7}
&0.15  &35.0\%  &63.3\%  &32.3\%  &\textbf{38.9\%} &35.3\%\\ \cline{2-7}
&0.20  &34.5\%  &62.7\%  &31.8\%  &\textbf{39.3\%} &35.0\%\\ \cline{2-7} 
&0.25  &34.7\%  &63.4\%  &31.7\%  &\textbf{39.1\%} &36.7\%\\ \cline{2-7} 
&0.30  &34.7\%  &61.9\%  &31.1\%  &\textbf{39.0\%} &36.3\%\\ 
\toprule
\multirow{7}{*}{4} 
&0     &78.9\%  &75.9\% &77.9\%  &77.0\%  &\textbf{78.9\%}\\  \cline{2-7}
&0.05  &42.2\%  &45.2\% &58.1\%  &58.8\% &\textbf{61.0\%}\\ \cline{2-7}
&0.10  &38.8\%  &46.2\% &56.1\%  &56.2\% &\textbf{61.0\%}\\ \cline{2-7}
&0.15  &37.9\%  &45.2\% &55.5\%  &54.7\% &\textbf{60.6\%}\\ \cline{2-7}
&0.20  &37.0\%  &44.2\% &55.0\% &54.3\% &\textbf{60.9\%}\\ \cline{2-7}
&0.25  &36.9\%  &44.7\% &54.8\% &54.0\% &\textbf{61.6\%}\\ \cline{2-7}
&0.30  &36.7\%  &44.7\% &54.5\%  &53.9\% &\textbf{60.1\%}\\ 
\bottomrule
\end{tabular}
\vspace*{-8pt}
\end{table}

\section{Conclusion}
\label{sec:conclusion}

In this paper, we discuss a new MTL model based on the VIB structure. The proposed framework has fundamental methodological contributions. First, the VIB structure is helpful to learn a compressive and low distorting latent representation of the input. Second, the task-dependent uncertainties are capable of learning the relative weights of the loss functions for the downstream tasks. The model is used to transform an MTL assignment into a constrained multi-objective optimization problem. Extensive results on three public data sets with the adversarial attacks at different levels show that the model has achieved comparable classification accuracy as the benchmark models. The observations suggest that our model can learn a more effective representation of the input and properly balance the weights of the downstream tasks. Our model has shown better robustness against adversarial attacks than the tested state-of-the-art MTL models. It is noted that other approaches such as mutual information neural estimation may be helpful to improve the estimation of mutual information. Additionally, semi-supervised learning methods are likewise applicable to learning the latent representations for multiple tasks. These issues will be explored in our future work.

\section*{Acknowledgment}
The first author acknowledges the China Scholarship Council with the funding support for this research. The second author would like to thank the funding support of the Region Bourgogne Franche Comt\'e Mobility Grant and the equipment support from Nvidia Corporation through its Accelerated Data Science Grant. Finally, the authors would also like to thank anonymous reviewers for their helpful comments on earlier drafts of the manuscript.

\bibliography{refs.bib}
\bibliographystyle{IEEEtran}


\begin{IEEEbiography}[{\includegraphics[width=1.1in,height=1.35in,clip,keepaspectratio]{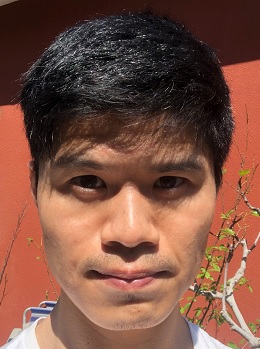}}]{Weizhu Qian}
is a PhD student in computer science at CIAD at Université de Technologie de Belfort-Montbéliard (UTBM) and Université Bourgogne-Franche-Comté (UBFC), France. He received his bachelor degree of science from Northwestern Polytechnical University, China in $2014$ and his master degree of science from University of Chinese Academy of Sciences, China in $2017$. His research interests include machine learning, deep learning, Bayesian statistics, variational inference, information theory and data science.\\
\vspace*{-10pt}
\end{IEEEbiography}

\begin{IEEEbiography}[{\includegraphics[width=1.1in,height=1.35in,clip,keepaspectratio,trim=0 0 0 0]{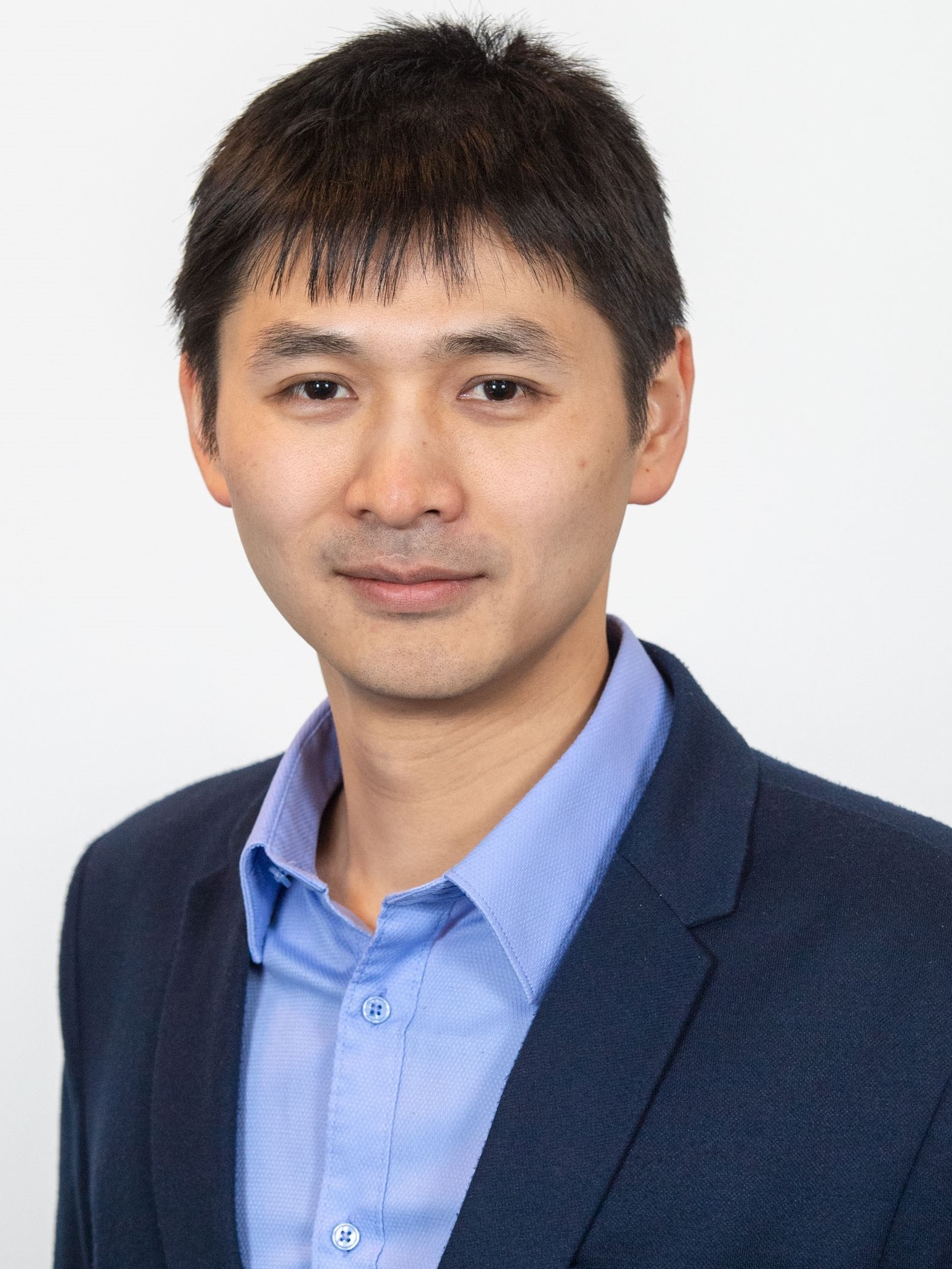}}]{Bowei Chen} is an Associate Professor at the Adam Smith Business School of University of Glasgow, UK. He received a PhD in Computer Science from University College London, and has broad research interests related to the applications of probabilistic modeling and deep learning in business, with special focuses on marketing and finance. He is on the editorial boards of Electronic Commerce Research and Applications, Frontiers in Big Data, and Frontiers in Artificial Intelligence. \\
\vspace*{-10pt}
\end{IEEEbiography}

\begin{IEEEbiography}[{\includegraphics[width=1.1in,height=1.35in,clip,keepaspectratio,trim=0 0 0 0]{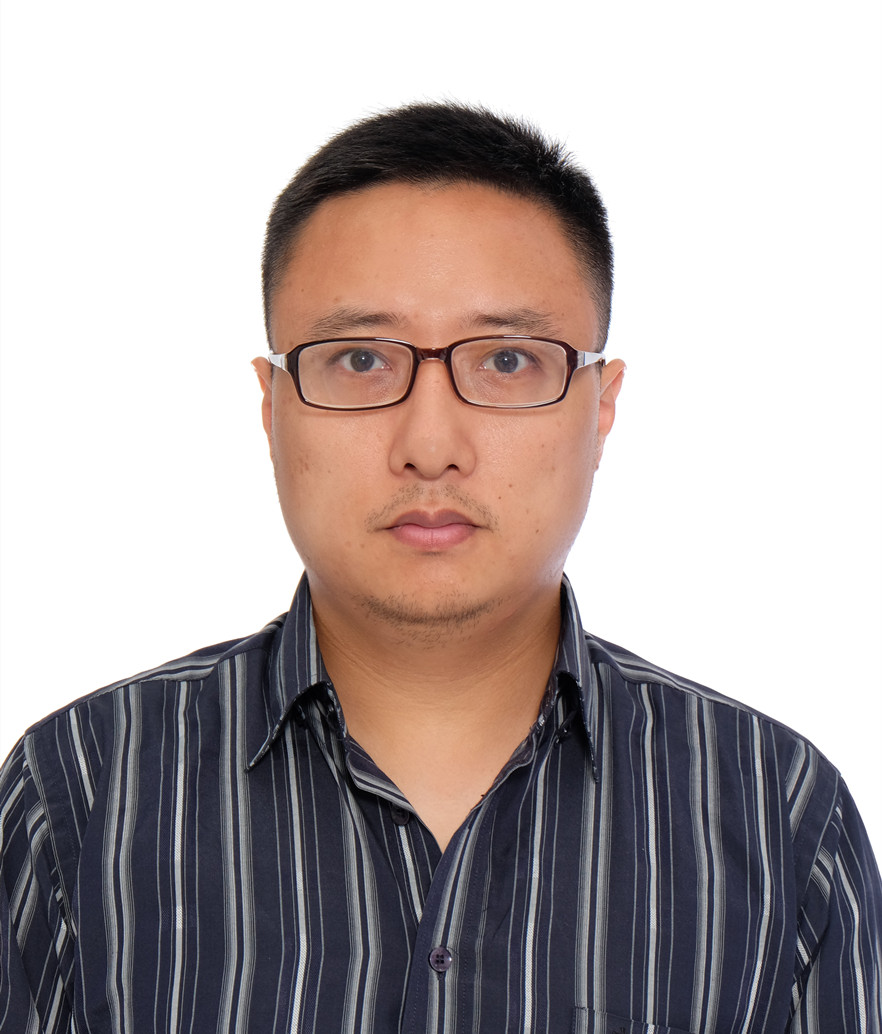}}]{Yichao Zhang} received the Ph.D degree in Computer Science and Technology from Tongji University, Shanghai, China. Currently, he is an Assistant Professor at the Department of Computer Science and Technology of Tongji University, Shanghai, China. His research interests include link prediction, modelling of weighted networks, random diffusion on weighted networks, and evolutionary games on networks.\\
\vspace*{-10pt}
\end{IEEEbiography}

\begin{IEEEbiography}[{\includegraphics[width=1.1in,height=1.35in,clip,keepaspectratio,trim=0 0 0 0]{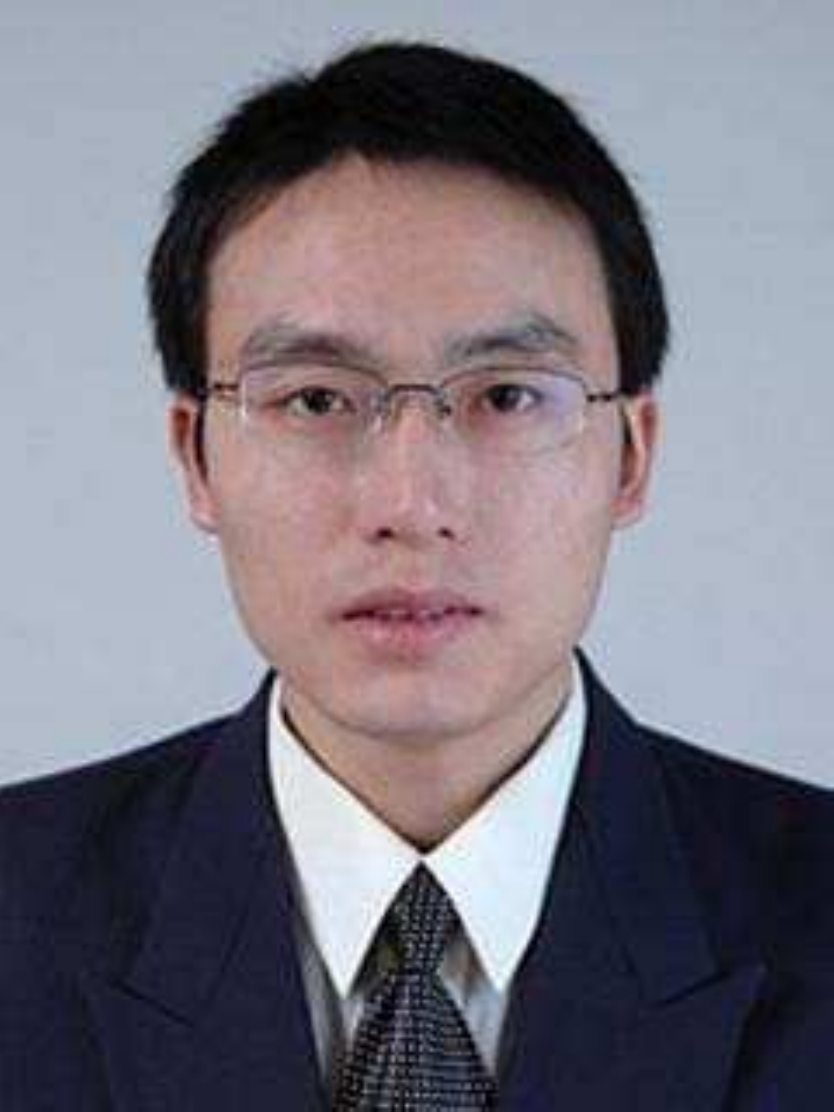}}]{Guanghui Wen} received his Ph.D. degree in mechanical systems and control from Peking University, China, in 2012.
Currently, he is a full professor with the Department of Systems Science, School of Mathematics, Southeast University, Nanjing, China. His research interests include cooperative control of multi-agent systems, analysis and synthesis of complex networks, cyber-physical systems, and resilient control.\\
\vspace*{-10pt}
\end{IEEEbiography}

\begin{IEEEbiography}[{\includegraphics[width=1.1in,height=1.35in,clip,keepaspectratio]{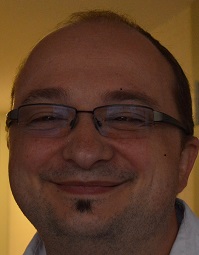}}]{Franck Gechter}
After a Master in engineering and a Master in photonics and image processing from the ULP Strasbourg I, F. Gechter received the Ph.D. in Computer Science from University H. Poincar\'{e} Nancy I (France) in 2003. In 2004, he became Associate Professor in Computer Science at UTBM and CIAD Laboratory UMR 7533 (Distributed Knowledge and Artificial Intelligence). In 2013, Franck Gechter passed his Habilitation to Lead Research Work (HDR) at the Franche Comté University (UFC). He works particularly on Reactive Multi-Agent models applied to problem solving, to decision processes and to data fusion. After focusing on issues related to Cyber-Physical systems control and simulation, he is now focusing on the Human/Cyber-Physical system interactions.\\
\vspace*{-10pt}
\end{IEEEbiography}

\end{document}